\documentclass[journal]{IEEEtran}
\usepackage{epsfig} 
\usepackage{mathptmx} 
\usepackage{times} 
\usepackage{amsmath} 
\usepackage{amssymb}  

\usepackage{multirow}
\usepackage{multicol}
\usepackage{soul}

\usepackage{graphicx}
\usepackage{threeparttable}

\usepackage{array}
\usepackage{subfigure}
\usepackage{leftidx}
\usepackage{float}
\usepackage{lineno}
\usepackage{hyperref}
\usepackage{array}
\usepackage{subfigure}
\usepackage{leftidx}
\usepackage{float}
\usepackage[dvipsnames]{xcolor}
\usepackage{amssymb}
\usepackage{pifont}
\newcommand{\cmark}{\ding{51}}%
\newcommand{\xmark}{\ding{55}}%

\ifCLASSINFOpdf
\else
\fi
\begin{document}
%
\title{Predicting Pedestrian Crossing Intention with Feature Fusion and Spatio-Temporal Attention}
%
%
%

\author {Dongfang Yang\textsuperscript{1,2,*}, Haolin Zhang\textsuperscript{1,*}, Ekim Yurtsever\textsuperscript{1}, \\
Keith Redmill\textsuperscript{1}, \IEEEmembership{Senior~Member, IEEE}, and \"{U}mit~\"{O}zg\"{u}ner\textsuperscript{1}, \IEEEmembership{Life~Fellow,~IEEE}
\thanks{*Equal contribution. Contact: yang.3455@osu.edu.}
\thanks{\textsuperscript{1}Department of Electrical and Computer Engineering, The Ohio State University, Columbus, OH 43210, USA. 
}
\thanks{\textsuperscript{2}Chang'an Automobile Co., Ltd., Chongqing, China.}

}
%
%

\markboth{Journal of \LaTeX\ Class Files,~Vol.~14, No.~8, August~2015}%
{Shell \MakeLowercase{\textit{et al.}}: Bare Demo of IEEEtran.cls for IEEE Journals}
%



\maketitle

\begin{abstract}
  Predicting vulnerable road user behavior is an essential prerequisite for deploying Automated Driving Systems (ADS) in the real-world. Pedestrian crossing intention should be recognized in real-time, especially for urban driving. Recent works have shown the potential of using vision-based deep neural network models for this task. However, these models are not robust and certain issues still need to be resolved. First, the global spatio-temproal context that accounts for the interaction between the target pedestrian and the scene has not been properly utilized. Second, the optimum strategy for fusing different sensor data has not been thoroughly investigated. This work addresses the above limitations by introducing a novel neural network architecture to fuse inherently different spatio-temporal features for pedestrian crossing intention prediction. We fuse different phenomena such as sequences of RGB imagery, semantic segmentation masks, and ego-vehicle speed in an optimum way using attention mechanisms and a stack of recurrent neural networks. The optimum architecture was obtained through exhaustive ablation and comparison studies.  Extensive comparative experiments on the JAAD pedestrian action prediction benchmark demonstrate the effectiveness of the proposed method, where state-of-the-art performance was achieved. Our code is open-source and publicly available: \url{https://github.com/OSU-Haolin/Pedestrian_Crossing_Intention_Prediction}. 
\end{abstract}


%
\IEEEpeerreviewmaketitle

\section{Introduction}

Autonmous driving technology has made significant progress in the past few years. However, to develop vehicle inteligence that is comparable to human drivers, understanding and predicting the behaviors of traffic agents is indispensable. This work aims to develop behavior understanding algorithms for vulnerable road users. Specifically, a vision-based pedestrian crossing intention prediction algorithm is proposed. 

Behavior understanding plays an crucial role in autonomous driving system. It establishes trust between people and autonomous dirving vehicles. By explicitly showing passengers how the system make its decisions, people will be more willing to accept this technology. 

In level 4 autnomous driving, pedestrian crossing behavior is one of the most important behaviors that needs to be studied urgently. In urban scenarios, vehicles frequently interact with crossing pedestrians. If the autonomous system failed to handle vehicle-pedestrian interaction, casualties will most likely occur. With accurate intention prediction, the decision-making and planning modules in autonomous driving systems can access additional meaningful information, hence generating more safe and efficient maneuvers.

Nowadays, visual sensors such as front-facing cameras are becoming the standard configuration of autonomous driving systems. In the tasks of object detection and tracking, both the software and hardware of vision components are mature and ready for mass production. This provides a perfect platform on which vision-based behavior prediction algorithms can be deployed. Researchers and engineers in the prediction field can just focus on algorithm design. When the algorithm is ready, deployment becomes relatively trivial. This makes the proposed algorithm, vision-based pedestrian intention prediction promising. As long as the prediction algorithm is appropriately tested and verified, mass deployment becomes straightforward.

\begin{figure}
  \centering
  \includegraphics[width=1\linewidth]{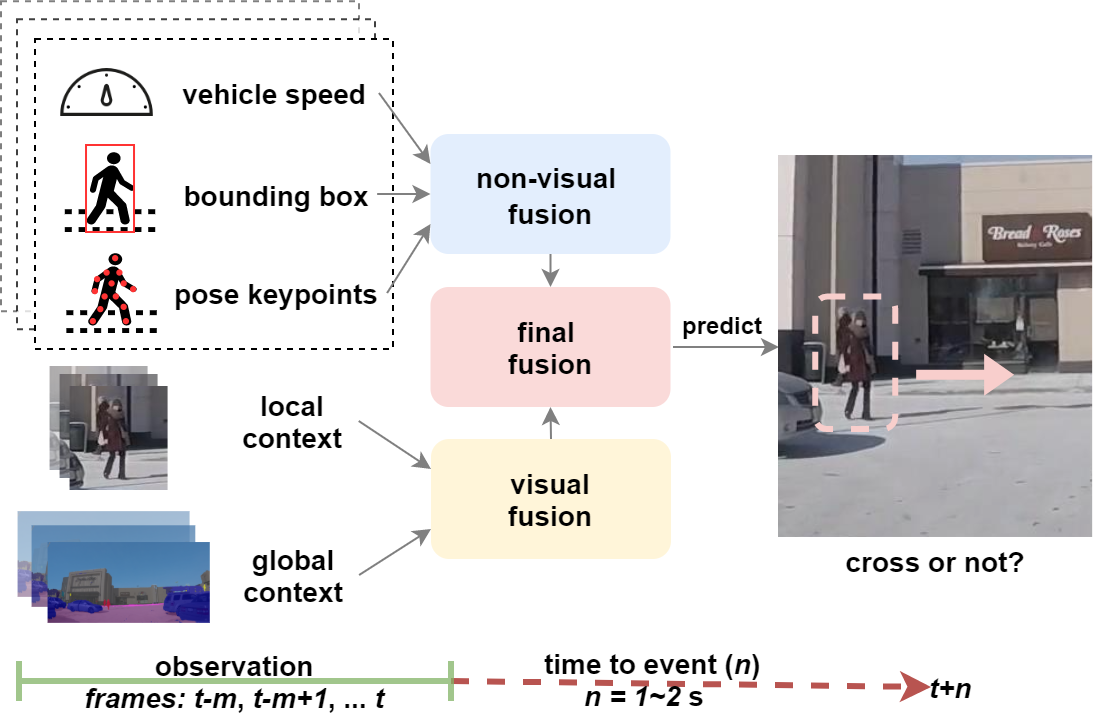}
  \vspace{0.1cm}
  \caption{Predicting pedestrian crossing intention is a multi-modal spatio-temporal problem. Our method fuses inherently different spatio-temporal phenomena with CNN-based visual encoders, RNN stacks, and attention mechanisms to achieve state-of-the-art performance. } 
  \label{fig:feature}
  \vspace{0.1cm}
\end{figure}

Vision-based pedestrian crossing intention prediction has been explored for several years. Early works~\cite{jaad} usually utilized a single frame as input to a convolutional neural network (CNN) based prediction system. This approach ignores the temporal aspect of image frames, which play a critical role in the intention prediction task. Later on, with the maturity of recurrent neural networks (RNNs), pedestrian crossing intention was predicted by considering both the spatial and temporal information \cite{singlernn,rnnbased,pie}. This led to different ways of fusing different features, e.g., the detected pedestrian bounding boxes, poses, appearance, and even the ego-vehicle information~\cite{pose1,fussi,sfrnn,multimodel,multitask}. The most recent benchmark of pedestrian intention prediction was released by \cite{pcpa}, in which the PCPA model achieved the state-of-the-art in the most popular dataset JAAD~\cite{jaad}. However, PCPA lacks the consideration of global contexts such as road and other road users. We believe they are nonnegligible in pedestrian crossing intention prediction. Furthermore, the existing fusion strategies may not be optimal.

In this work, we focus on improving the performance of vision-based prediction of pedestrian crossing intention, i.e., whether a pedestrian detected by a front-facing camera will cross the road or not in a short time horizon (1-2s). Our work leverages the power of deep neural networks and fuses the features from different channels. As shown in Figure~\ref{fig:feature}, the propsoed model considers both non-visual and visual information. They are extracted from a sequence of video frames 1-2s before the crossing / not crossing (C/NC) event. Non-visual information includes the pedestrian's bounding box, pose keypoints, and ego-vehicle speed. Visual information contains local context and global context. Local context is the enlarged pedestrian appearance based on the bounding box position. Global context is the semantic segmentation of road, pedestrians (all pedestrians in the scene), and vehicles. They are used because they significantly affect the target pedestrian's crossing decision. We proposed a hybrid way of fusing the the non-visual and visual features, which is justified by comparing different strategies of feature fusion.



  Our main contributions are as follows:
  \begin{itemize} 

    \item A novel vision-based pedestrian intention prediction framework for ADSs and ADASs. The proposed method employs a novel neural network architecture for utilizing different spatio-temporal features with a hybrid fusion strategy. 

    \item Extensive ablation studies on different feature fusion strategies (early, later, hierarchical, or hybrid), input configurations (adding/removing input channels, using semantic segmentation masks as explicit global context), and visual encoder options (3D CNN or 2D convolution with RNN + attention) to identify the best model layout.

    \item Demonstrating the efficiency of the proposed method on the commonly used JAAD dataset~\cite{jaad}, and achieving state-of-the-art performance on the most recent pedestrian action prediction benchmark~\cite{pcpa}. 
    
    
      
  \end{itemize}

\section{Related Work}

Vision-based pedestrian crossing prediction traces back to the works~\cite{early} that utilize the Caltech Pedestrian Detection Benchmark~\cite{detection}. However, the Caltech dataset does not explicitly annotate the crossing behavior of the pedestrians. This gap was later filled by the introduction of JAAD dataset~\cite{jaad} that offers high-resolution videos and explicit crossing behavior annotations. With the release of JAAD dataset, a simple baseline was also created that uses a 2D convolutional neural network (CNN) to encode the features in a given previous frame and then uses a linear support vector machine (SVM) to predict the C/NC event. 


\textbf{Spatio-temporal modeling.} Instead of using a single image, most recent works use image sequences as input to the prediction model due to the importance of temporal information in the prediction task. This leads to spatio-temporal modeling. 

Spatio-temporal modeling can be achieved by first extracting visual (spatial) features per frame via 2D CNNs~\cite{vgg} or graph convolution networks (GCNs)~\cite{gcns}, and then feeding these features into RNNs such as long-short term memory (LSTM) model~\cite{LSTM} and gate recurrent unit (GRU) model~\cite{GRU}. For example, \cite{singlernn,rnnbased,pie} use 2D convolution to extract the visual features from image sequence, and RNNs to encode the temporal information among these features. The encoded sequential visual features are fed into a fully-connected layer to obtain the final intention prediction.

Another way of extracting the sequential visual features is utilizing 3D CNN~\cite{3dcnn}. It directly captures the spatio-temporal features by replacing the 2D kernels of the convolution and the pooling layers in 2D CNN with 3D counterparts. For example, ~\cite{densenet,densenet2} use 3D CNN based framework (3D DenseNet) to directly extract the sequential visual features from the pedestrian image sequence. The final prediction is achieved in a similar way of using a fully-connected layer.

The crossing intention prediction task can also be combined with scene prediction. A couple of works~\cite{frame1,frame2} attempted to decompose the prediction task into two stages. In the first stage, the model predicts a sequence of future scenes using an encoder/decoder network. Then, pedestrian actions are predicted based on the generated future scenes using a binary classifier.



\begin{figure*}[htb]
  \begin{center}{\includegraphics[width=1\linewidth]{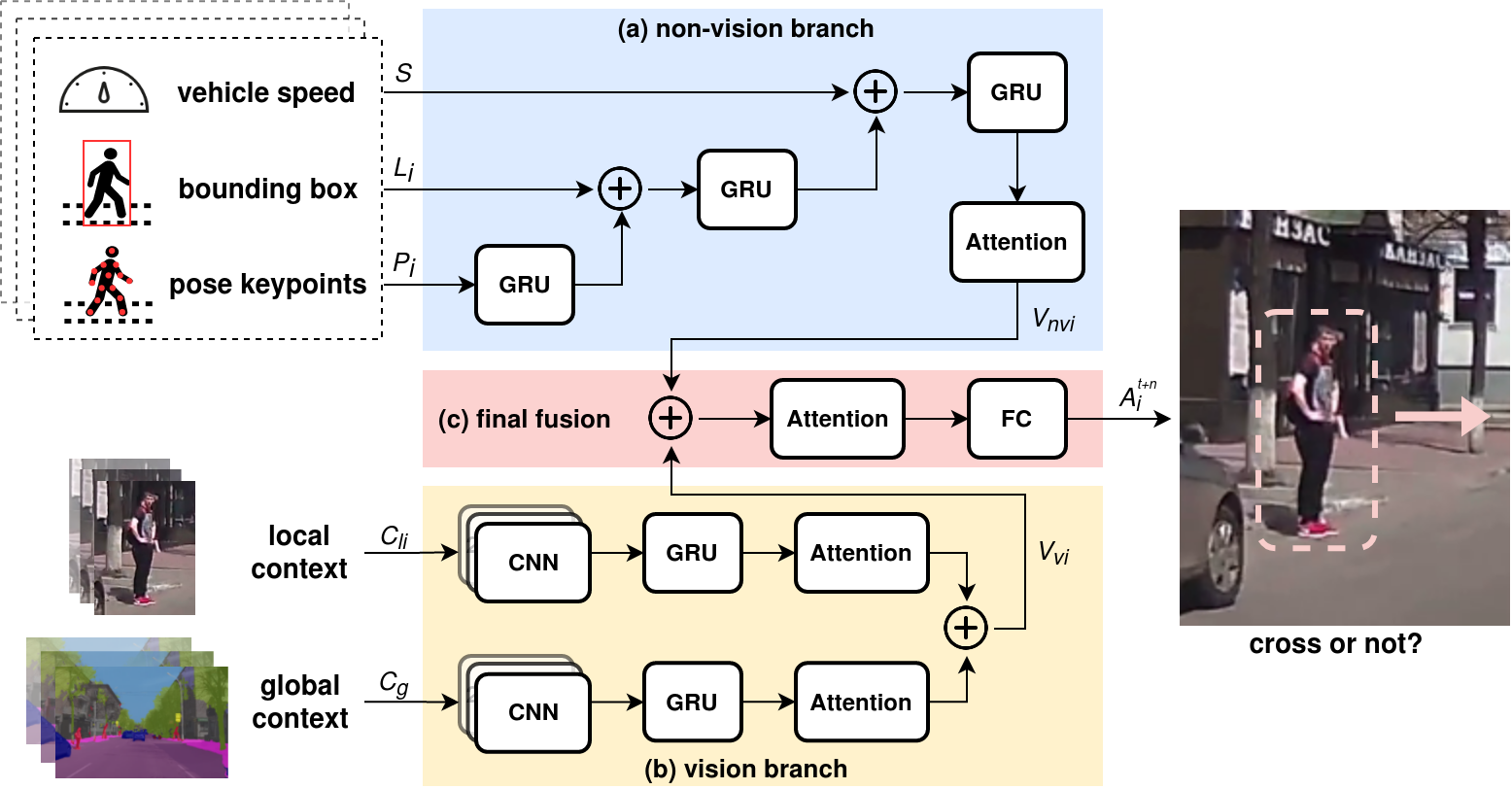}}
  \end{center}
  \vspace{0.25cm}
    \caption{{\bf Overview of the proposed pedestrian crossing intention prediction model.} The yellow part denotes the fusion of visual features. 2D convolutional features of local context and global context are encoded by GRU and fed to the attention blocks respectively. Two outputs are concatenated as final visual features. The blue part denotes the fusion of local features (non-visual). These non-visual features are encoded by GRU and fused hierarchically, and then fed to an attention block to obtain the final non-visual features. The red part denotes the final fusion. Final visual features and final non-visual features are concatenated and fed to an attention block. A fully-connected (FC) layer is then applied to make the final prediction.} 
    \vspace{0.25cm}
  \label{fig:pipeline}
  \end{figure*}


\textbf{Feature fusion.} Instead of end-to-end modeling of visual features, information such as pedestrian's bounding box, body-pose keypoints, vehicle motion, and the explicit global scene context can also be modeled as separate channels as inputs to the prediction model. This requires a proper way of fusing the above information. 

For example, \cite{pose1,pose2,pose3,pose4,fussi} introduced human poses/skeletons in pedestrian crossing prediction tasks since human pose can be considered as a good indicator of human behaviors. By extracting the pose keypoints from cropped pedestrian images, crossing behavior classifiers are built based on the human pose feature vectors. Improvement in prediction accuracy shows the effectiveness of using pose features. However, these methods either only rely on human pose features without considering other important features or pay less attention to feature fusion.   

Some other methods focused on novel fusion architecture. For instance, \cite{sfrnn} proposed SF-GRU, a stacked RNN-based architecture, to hierarchically fuse five feature sources (pedestrian appearance, surrounding context, pose, bounding box, and ego-vehicle speed) for pedestrian crossing intention prediction. Nevertheless, this method does not take global context into account. \cite{multimodel} proposed a multi-modal based prediction system that integrates four feature sources (local scene, semantic map, pedestrian motion, and ego-motion). The global context (semantic map) is utilized, but it lacks other important features such as human pose. \cite{multitask} proposed a multi-task based prediction framework to take advantages of feature sharing and multi-task learning. It fuses four feature sources (semantic map, pedestrians' trajectory, grid locations, and ego-motion). However, local context and human pose are not considered in the model.

Very recently, more datasets such as PIE~\cite{pie} and PePScenes~\cite{pep} provide more annotations for fusing different features. A benchmark was also released with the PCPA model~\cite{pcpa}. They create more room for researchers to explore the task of vision-based pedestrian crossing intention prediction.






\section{Proposed Method}

\subsection{Problem formulation}

The task of vision-based pedestrian crossing intention prediction is formulated as follows. Given a sequence of observed video frames from the vehicle's front view and the relevant information of ego-vehicle motion, the goal is to design a model that can estimate the probability of the target pedestrian $i$'s action $A^{t+n}_{i}$ $\in \{0,1\}$ of crossing the road, where $t$ is the specific time of the last observed frame and $n$ is the number of frames from the last observed frame to the crossing / not crossing (C/NC) event.


In the proposed model, explicit features such as pedestrian's bounding box, pose keypoints, local context (cropped image around the pedestrian), and global context (semantic segmentation) are firstly extracted. They are used together with the vehicle's speed as separate channels that serve as the input to the prediction model. Therefore, our model has the following input sources:

\begin{itemize}
  \item The sequential local context around pedestrian $i$:
  \begin{equation}
    C_{li} = \{ c^{t-m}_{li}, c^{t-m+1}_{li},  ... ,c^{t}_{li} \}; \notag
  \end{equation}
  \item The 2D location trajectory of pedestrian $i$ denoted by bounding box coordinates (top-left points and bottom-right points): 
  \begin{equation}
    L_{i} =\{ l^{t-m}_{i}, l^{t-m+1}_{i},  ... ,l^{t}_{i} \} ; \notag
  \end{equation}
  \item Pose keypoints of pedestrian $i$: 
  \begin{equation}
    P_{i} =\{ p^{t-m}_{i}, p^{t-m+1}_{i}, ... ,p^{t}_{i} \}; \notag
  \end{equation}
  \item Speed of ego-vehicle: 
  \begin{equation}
    S = \{ s^{t-m}, s^{t-m+1}, ... ,s^{t} \} ; \notag
  \end{equation}

  \item The sequential global context denoted by the mask of semantic segmentation:
  \begin{equation}
    C_g =\{ c^{t-m}_{g}, c^{t-m+1}_{g}, ... ,c^{t}_{g} \}. \notag
  \end{equation}

\end{itemize}

Each source has a sequence of length $m+1$. The input sources are illustrated in Figure \ref{fig:pipeline}.


\subsection{Input acquisition}


{\bf Local context and 2D location trajectory}. Local context $C_{li}$ provides visual features of the target pedestrian. 2D location trajectory $L_{i}$ gives the position change of the target pedestrian in the image. They can be extracted by a detection (e.g. YOLO~\cite{yolo}) and tracking (e.g. SORT~\cite{tracking}) system. In our work, we directly use the ground truth $C_{li}$ and $L_{i}$ from the dataset, because pedestrian detection and tracking are not the primary focus of this work. Specifically, the local context $C_{li} = \{ c^{t-m}_{li}, c^{t-m+1}_{li},  ... ,c^{t}_{li} \} $ consists of a sequence of RGB images of size $[224,224]$ pixels around the target pedestrian. The 2D location trajectory $L_{i} =\{ l^{t-m}_{i}, l^{t-m+1}_{i},  ... ,l^{t}_{i} \} $ consists of target pedestrian's bounding box coordinates, i.e., 
\begin{equation}
  l^{t-m}_{i} = \{ x^{t-m}_{it}, y^{t-m}_{it}, x^{t-m}_{ib}, y^{t-m}_{ib}\}, \notag
\end{equation}
where $x^{t-m}_{it}, y^{t-m}_{it}$ denotes the top-left point and $x^{t-m}_{ib}, y^{t-m}_{ib}$ bottom-right point.

{\bf Pedestrian pose keypoints}. Pedestrian pose keypoints represent the target pedestrian's detailed motion, i.e., the posture at each frame while moving. They can be obtained by applying a pose estimation algorithm on the local context $C_{li}$. Since the applied JAAD dataset does not provide ground truth pose keypoints, we utilize pre-trained OpenPose model \cite{openpose} to extract the pedestrian pose keypoints $P_{i} =\{ p^{t-m}_{i}, p^{t-m+1}_{i}, ... ,p^{t}_{i} \} $, where $p$ is a 36D vector of 2D coordinates that contain 18 pose joints, i.e., 
\begin{equation}
  p^{t-m}_{i} = \{ x^{t-m}_{i1}, y^{t-m}_{i1}, x^{t-m}_{i2}, y^{t-m}_{i2}, ... ,x^{t-m}_{i18}, y^{t-m}_{i18} \}. \notag
\end{equation}




{\bf Ego-vehicle speed.} Ego-vehicle speed $S$ is a major factor that affects the pedestrian's crossing decision. It can be directly read from the ego-vehicle's system. Since the dataset contains the annotation of ego-vehicle's speed, we directly use the ground truth labels for the vehicle speed $S = \{ s^{t-m}, s^{t-m+1}, ... ,s^{t} \} $.


{\bf Global context.} Global context $C_g = \{ c^{t-m}_{g}, $ $c^{t-m+1}_{g}, ... ,c^{t}_{g} \} $ offers the visual features that account for multi-interactions between the road and road users, or among road users. In our work, we use pixel-level semantic masks to represent the global context. The semantic masks classify and localize different objects in the image by labeling all the pixels associated with the objects the a pixel value.
Since the JAAD dataset does not have annotated ground truth of semantic masks, we use DeepLabV3 model \cite{deeplabv3} pretrained on Cityscapes Dataset \cite{city} to extract the semantic masks and select important objects (e.g. road, street, pedestrians and vehicles) as the global context.
For the model to learn the interactions between the target pedestrian $i$ and these objects, the target pedestrian is masked by an unique label. The mask area uses the target pedestrian $i$'s bounding box (obtained from $L_{i}$).
The semantic segmentation of all input frames are scaled to be the size of $[224,224]$ pixels, which is the same as the local context. 






\subsection{Model architecture}

The overall architecture is shown in Figure~\ref{fig:pipeline}. It consists of CNN modules, RNN modules, attention modules, and a novel way of fusing different features.

{\bf CNN module.} 
We use VGG19 \cite{vgg} model pre-trained on ImageNet dataset \cite{imagenet} to build the CNN module. Sequential RGB images are collected as a 4D array input with the dimensions of [number of observed frames, row, cols, channels] ($[16, 224, 224, 3]$ in this work), and then loaded by the CNN module. First, the feature map of every image from the fourth maxpooling layer of VGG19 is extracted with size $[512,14,14]$. Second, every feature map is averaged by a pooling layer with a $14 \times 14$ kernel, and then flattened and concatenated, to obtain a final feature tensor with size $[16,512]$, as sequential visual features.

{\bf RNN module.} 
We use gated recurrent unit (GRU) \cite{GRU} to build the RNN module. The reason of choosing GRU is that GRU is more computationally efficient than its counterpart LSTM~\cite{LSTM}, which is older, and its architecture is relatively simple. 
The applied GRUs have 256 hidden units, which result in a feature tensor of size $[16,256]$





{\bf Attention module.} Attention module~\cite{attention}, by selectively focusing on parts of features, is used for better memorizing sequential sources. 
Sequential features (e.g. the output of RNN-based encoder) are represented as hidden states $h = \{ h_{1}, h_{2}, ..., h_{e}\}$. The attention weight is computed as: 
\begin{equation}
  \alpha = \frac{exp(score(h_{e},\tilde{h_{s}}))}{\sum_{s^{\prime}}exp(score(h_{e},\tilde{h_{s^{\prime}}}))},  \notag
\end{equation}
where $score(h_{e},\tilde{h_{s}}) = h_{e}^{T}W_{s}\tilde{h_{s}}$ and $W_{s}$ is a weight matrix. Such attention weight trades off the end hidden state $h_{e}$ with each previous source hidden state $\tilde{h_{s}}$. The output vector of the attention module is produced as 
\begin{equation}
  V_{attention} = tanh(W_{c}[h_{c};h_{e}]), \notag
\end{equation}
where $W_{c}$ is a weight matrix, and $h_{c}$ is the sum of all attention weighted hidden states as $h_{c} = {\sum_{s^{\prime}}}{\alpha}{\tilde{h_{s^{\prime}}}}$. The output of the attention module in our work is a feature tensor with size $[1,256]$.



{\bf Hybrid fusion.} We applied a hybrid way of fusing the features from different sources. The strategy is shown in Figure \ref{fig:pipeline}. The proposed architecture has two branches, one for non-visual features and one for visual features. 

The non-vision branch fuses three non-visual features (bounding boxes, pose keypoints, and vehicle speed). They are hierarchically fused according to their complexity and level of abstraction. The later stage of fusion, the closer impact of the fused feature on final prediction. This is illustrated in Figure~\ref{fig:pipeline}(a). 
First, sequential pedestrian pose keypoints $P_{i}$ are fed to a RNN-based encoder. Second, the output of the first stage is concatenated with 2D location trajectory $L_{i}$ and fed to a new RNN-based encoder. Last, the output of the second stage is concatenated with ego-vehicle speed $S$ and fed to a final RNN-based encoder. The output of the final encoder is then fed to an attention block to obtian the final non-visual feature vectors $V_{nvi}$. 





The vision branch fuses two visual features, consisting of local context (enlarged pedestrian appearance around the bounding box) and global context (semantic segmentation of important objects in the whole scene), as shown in Figure~\ref{fig:pipeline}(b). Local context $C_{li}$ is encoded by first extracting spatial features from the CNN module (as explained in the previous section) and then extracting temporal features from the GRU module.
Global context $C_{g}$ is encoded in the same way. 
Both local and global features are then fed into their attention modules, and finally, concatenated together to generate final visual feature vectors $V_{vi}$.

\begin{figure}[h]
  \includegraphics[width=1\linewidth]{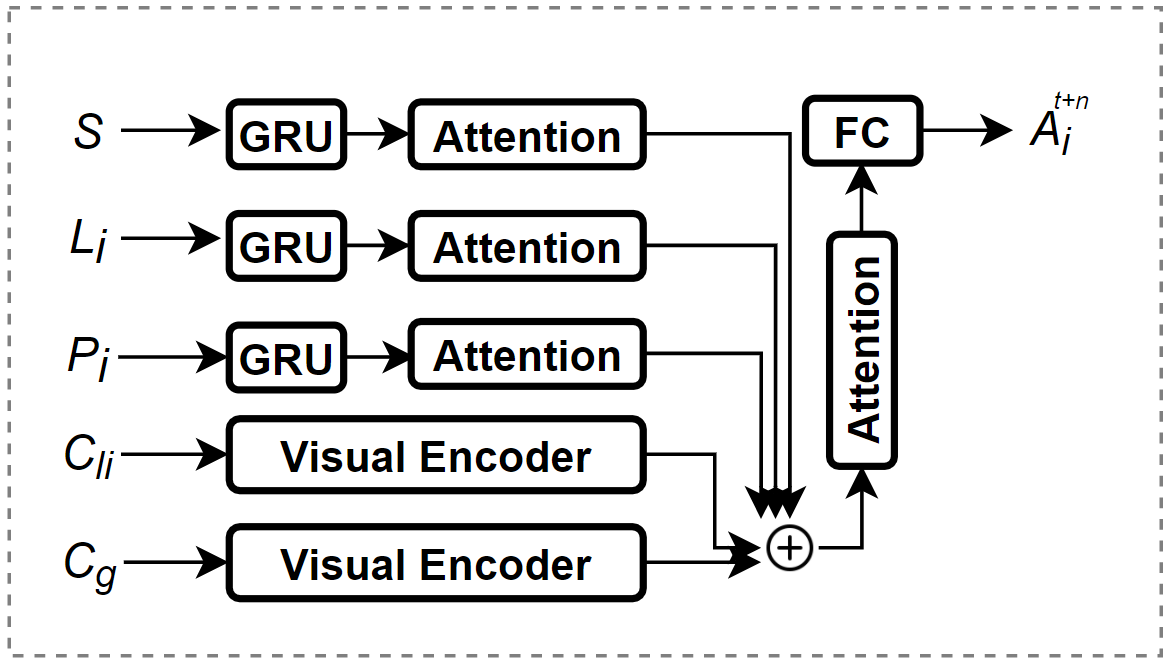}
  \caption{Illustration of Later Fusion}
  \label{fig:fusion_later}
\end{figure}

\begin{figure}[h]
  \includegraphics[width=1\linewidth]{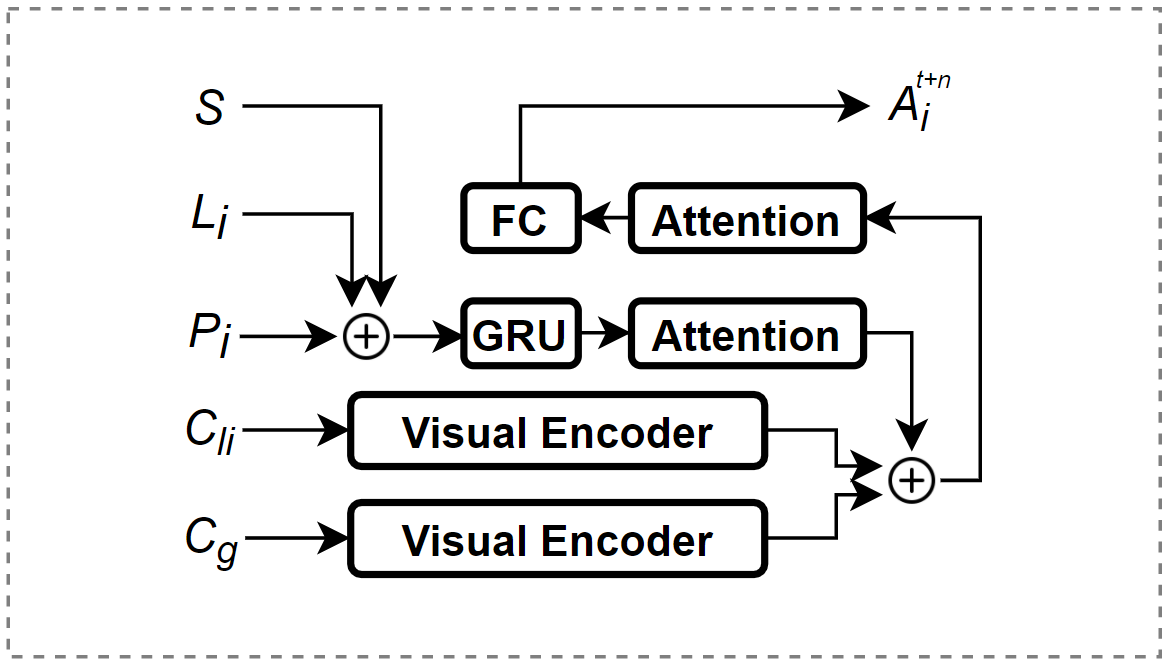}
  \caption{Illustration of Early Fusion}
  \label{fig:fusion_early}
\end{figure}

\begin{figure}[h]
  \includegraphics[width=1\linewidth]{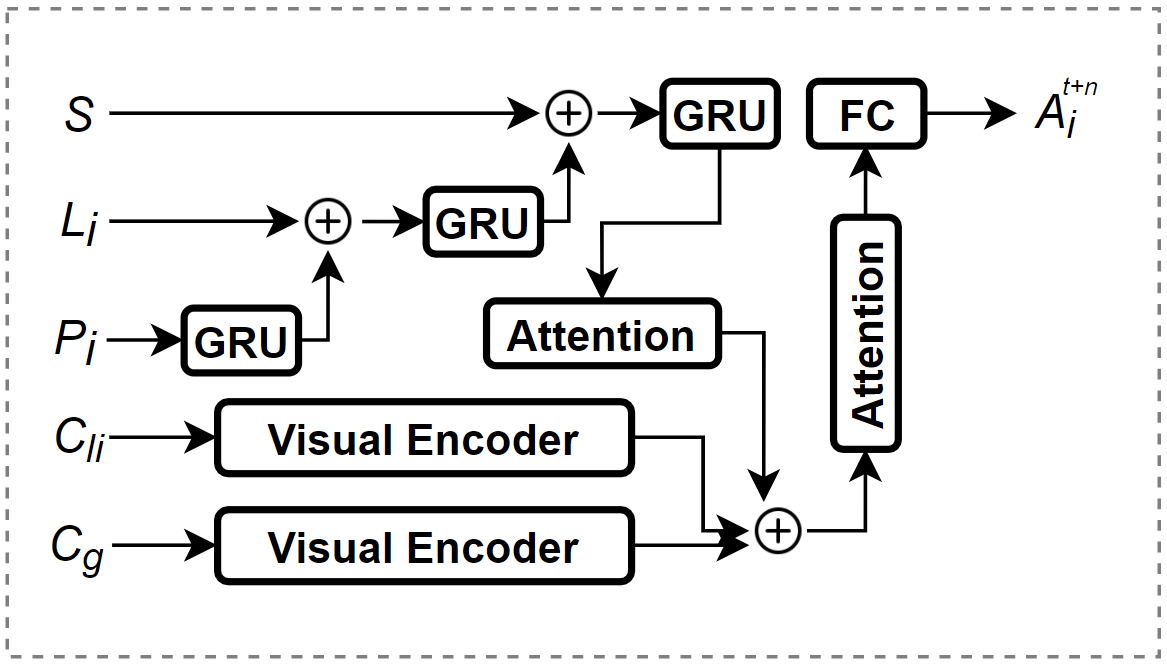}
  \caption{Illustration of Hierarchical Fusion}
  \label{fig:fusion_hierarchical}
\end{figure}

Lastly, as shown in Figure~\ref{fig:pipeline}(c), the final non-visual feature vectors $V_{nvi}$ and the final visual feature vectors $V_{vi}$ are concatenated and fed into another attention block, followed by a fully-connection (FC) layer to obtain the final predicted action:
\begin{equation}
  A_{i}^{t+n} = f_{FC}(f_{attention}(V_{nvi};V_{vi})). 
  \notag
\end{equation}

\section{Experiments} 

\subsection{Dataset and Benchmark}

The proposed model was evaluated using JAAD dataset \cite{jaad}. It contains two subsets, JAAD behavioral data (JAAD$_{beh}$) and JAAD all data (JAAD$_{all}$). JAAD$_{beh}$ contains pedestrians who are crossing (495 samples) or are about to cross (191 samples). JAAD$_{all}$ has additional pedestrians (2100 samples) with non-crossing actions. To create a fair benchmark, the dataset configuration follows the same one as in~\cite{pcpa}. It uses a data sample overlap of 0.8, local context scale of 1.5. The evaluation metrics use accuracy, AUC, F1 score, precision, and recall. They are the most recognized metrics and are used by most related works. 


\subsection{Implementation}

In the expriments, the propsoed model was compared with the following methods: SingleRNN \cite{singlernn}, SF-GRU \cite{sfrnn} and PCPA\cite{pcpa}. We adopted the benchmark implementation released with PCPA model~\cite{pcpa}. This benchmark collects the implementations of most pedestrian intention prediction methods. Our model was developed based on the this benchmark. We use a dropout of 0.5 in the attention module, L2 regularization of 0.001 in FC layer, binary cross-entropy loss, Adam optimizer \cite{adam}, learning rate = $5 \times 10^{-7}$, epochs = 40, and batch size = 2. All models were trained and tested on the same split of the dataset, as suggested by the benchmark. Note that JAAD dataset does not provide explicit vehicle speed. Instead, the driver's action is recorded as an abstract encoding of the vehicle speed. The action contains $[$stopped (0), moving slow (1), moving fast (2), decelerating (3), accelerating (4)$]$.


\subsection{Ablation study}

An ablation study was also conducted to compare different strategies of fusing different features.
In addition to baseline methods (SingRNN~\cite{singlernn}, SF-GRU~\cite{sfrnn} and PCPA~\cite{pcpa}) and the proposed model (Ours), a total of 7 variants of the proposed model (Ours1, Ours2, ... , Ours7, as indicated in table~\ref{table:jad_beh_ablation} and table~\ref{table:jad_all_ablation}) were trained and compared with the proposed one. 
First, for the visual encoder, we tried (1) 2D CNN combined with RNN (VGG and GRU in our experiments) and (2) 3D CNN as proposed in the PCPA model. 
Second, we tried the models with and without the global feature (semantic segmentation). 
Last, we tried different fusion strategies that include later fusion, early fusion, and hierarchical fusion so that they can be compared with the proposed hybrid fusion strategy. Later fusion (Figure~\ref{fig:fusion_later}) is the same as the one proposed in PCPA~\cite{pcpa}. Early fusion (Figure~\ref{fig:fusion_early}) concatenates non-visual features and visual features directly and then send them into one RNN module followed by an attention module. Hierarchical fusion (Figure~\ref{fig:fusion_hierarchical}) gradually fuses both visual features and non-visual features by RNN modules using the same way as in Figure~\ref{fig:pipeline}(a), followed by an attention module.



\section{Results}

\begin{figure*}[ht]
  \includegraphics[width=1\linewidth]{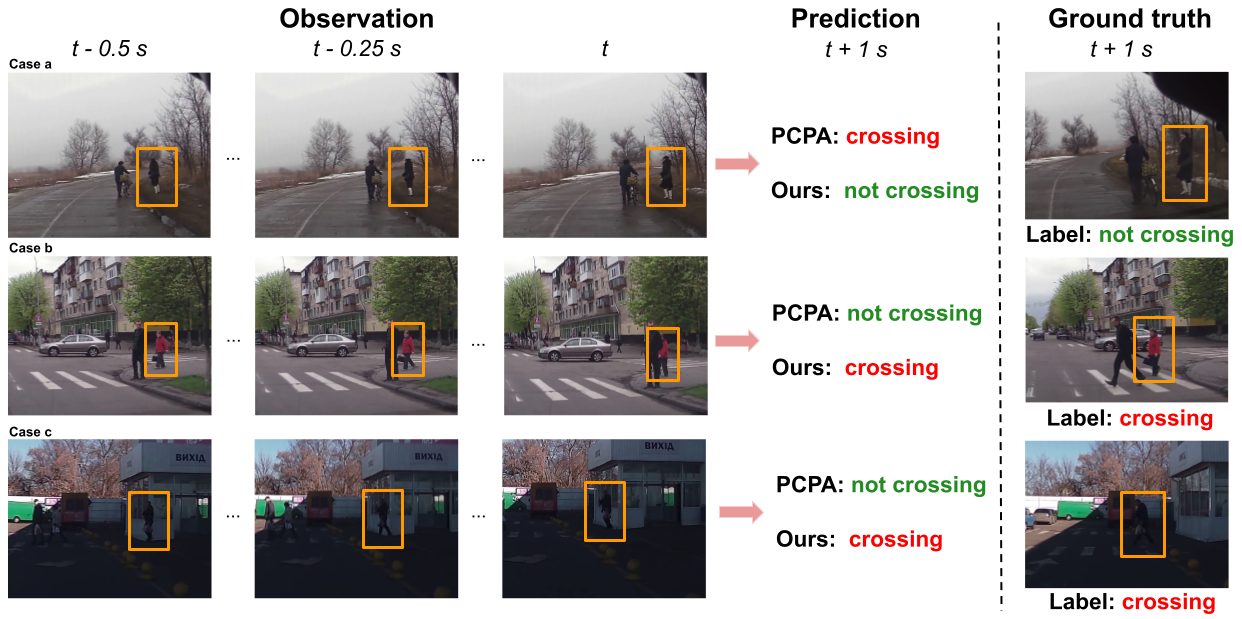}
  \vspace{0.25cm}
  \caption{{\bf Qualitative results on the JAAD dataset} produced by PCPA~\cite{pcpa} and our proposed model (Ours). The target pedestrians in images are enclosed by {\bf \textcolor{BurntOrange}{bounding boxes}}. The prediction results as well as ground truth labels are represented as  {\bf \textcolor[rgb]{1,0,0}{crossing}} or {\bf \textcolor{ForestGreen}{not crossing}}.} 
  \label{fig:results}
  \vspace{0.25cm}
\end{figure*}

\begin{table*}[htb]
  \centering
  \centering
  \caption{Quantitative Results on JAAD Behavior Subset}
  \begin{threeparttable}
  \small
  \renewcommand{\arraystretch}{1.3}
  
  \begin{tabular}{ccccccccc}
  
  \hline
  \hline
  
  \multirow{2}{*}{\bf Models}
  & \multicolumn{3}{|c|}{\begin{tabular}[c]{@{}c@{}} {\bf Model Variants} \end{tabular}} 
  & \multicolumn{5}{c}{\begin{tabular}[c]{@{}c@{}} {\bf JAAD$_{beh}$} \end{tabular}} 
  \\ \cline{2-9} 

  & \multicolumn{1}{|c|}{\begin{tabular}[c]{@{}c@{}} Visual Encoder \end{tabular}}
  & \multicolumn{1}{c|}{\begin{tabular}[c]{@{}c@{}} Global Context \end{tabular}} 
  & \multicolumn{1}{c|}{\begin{tabular}[c]{@{}c@{}} Fusion Approach\end{tabular}} 
  
  & \multicolumn{1}{c|}{\begin{tabular}[c]{@{}c@{}} Accuracy \end{tabular}} 
  & \multicolumn{1}{c|}{\begin{tabular}[c]{@{}c@{}} AUC \end{tabular}} 
  & \multicolumn{1}{c|}{\begin{tabular}[c]{@{}c@{}} F1 Score \end{tabular}} 
  & \multicolumn{1}{c|}{\begin{tabular}[c]{@{}c@{}} Precision \end{tabular}} 
  & \multicolumn{1}{c}{\begin{tabular}[c]{@{}c@{}} Recall \end{tabular}} 
  
   \\ \cline{1-9} 
  
  SingleRNN~\cite{singlernn} &VGG + GRU &\xmark &\xmark &0.59 &0.52 &0.71 &0.64 &0.80 \\
  SF-GRU~\cite{sfrnn} &VGG + GRU &\xmark &hierarchical-fusion &0.58 &\bf 0.56 &0.65 &{\bf 0.68} &0.62 \\
  PCPA~\cite{pcpa}      &3D CNN &\xmark &later-fusion &0.53 &0.53 &0.59 &0.66 &0.53  \\ \hline  
  Ours     &VGG + GRU &\cmark &hybrid-fusion &\bf0.62 &0.54 &{\bf0.74} &0.65 &\bf0.85 
  \\ \hline \hline
  
  \hline

  \end{tabular}

  \begin{tablenotes}
          \footnotesize
          \item[$\bullet$] The {\bf bold} result means the best in the models. 
        \end{tablenotes}
      \end{threeparttable}
  
  \label{table:jad_beh_comp}
  \end{table*}

\begin{table*}[htb]
  \centering
  \centering
  \caption{Quantitative Results on JAAD All Dataset}
  \begin{threeparttable}
  \small
  \renewcommand{\arraystretch}{1.3}
  
  \begin{tabular}{ccccccccc}
  
  \hline
  
  \hline
  
  \multirow{2}{*}{\bf Models}
  & \multicolumn{3}{|c|}{\begin{tabular}[c]{@{}c@{}} {\bf Model Variants} \end{tabular}} 
  & \multicolumn{5}{c}{\begin{tabular}[c]{@{}c@{}} {\bf JAAD$_{all}$} \end{tabular}} 
  \\ \cline{2-9} 

  & \multicolumn{1}{|c|}{\begin{tabular}[c]{@{}c@{}} Visual Encoder \end{tabular}}
  & \multicolumn{1}{c|}{\begin{tabular}[c]{@{}c@{}} Global Context \end{tabular}} 
  & \multicolumn{1}{c|}{\begin{tabular}[c]{@{}c@{}} Fusion Approach\end{tabular}} 
  
  & \multicolumn{1}{c|}{\begin{tabular}[c]{@{}c@{}} Accuracy \end{tabular}} 
  & \multicolumn{1}{c|}{\begin{tabular}[c]{@{}c@{}} AUC \end{tabular}} 
  & \multicolumn{1}{c|}{\begin{tabular}[c]{@{}c@{}} F1 Score \end{tabular}} 
  & \multicolumn{1}{c|}{\begin{tabular}[c]{@{}c@{}} Precision \end{tabular}} 
  & \multicolumn{1}{c}{\begin{tabular}[c]{@{}c@{}} Recall \end{tabular}} 
  
   \\ \cline{1-9}

  SingleRNN~\cite{singlernn} &VGG + GRU &\xmark &\xmark &0.79 &0.76 &0.54 &0.44 &0.71 \\
  
  SF-GRU~\cite{sfrnn} &VGG + GRU &\xmark &hierarchical-fusion  &0.76 &0.77 &0.53 &0.40 &0.79 \\
  
  PCPA~\cite{pcpa}      &3D CNN &\xmark &later-fusion &0.76 &0.79 &0.55 &0.41 &\bf 0.83 \\ \hline
  
  Ours     &VGG + GRU &\cmark &hybrid-fusion &\bf0.83 &\bf0.82 &\bf0.63 &\bf0.51 &0.81  \\ \hline

  \hline
  
  \end{tabular}

  \begin{tablenotes}
          \footnotesize
          \item[$\bullet$] The {\bf bold} result means the best in the models. 
        \end{tablenotes}
      \end{threeparttable}
  
  \label{table:jad_all_comp}
  \end{table*}

\subsection{Quantitative Results}

Table~\ref{table:jad_beh_comp} shows the qualitative results on JAAD$_{beh}$ dataset. It compares the proposed model with baseline models of SingleRNN~\cite{singlernn}, SF-GRU~\cite{sfrnn} and PCPA~\cite{pcpa}. The proposed model achieved the best scores in accuracy, F1, and recall. F1 score is an balanced metirc considering both recall and percision. For binary classification, it is the most important indicator of how good the model is. Our model achieved about 4\% improvement in F1. In addition to F1, accuracy is another important metric. Our model also achieved the best.

Table~\ref{table:jad_all_comp} shows the qualitative results on JAAD$_{all}$ dataset. JAAD$_{all}$ has additional samples of non-crossing behaviors. It is larger than JAAD$_{beh}$. The data distribution is more similar to real world scenarios. As illustrated by table~\ref{table:jad_all_comp}, the proposed achieved the best in accuracy, AUC, F1, precision. Similar to the results in JAAD$_{beh}$, our model achieved the best in terms of the two important metrics, F1 and accuracy. 

Table~\ref{table:jad_beh_ablation} and table~\ref{table:jad_all_ablation} show the results of ablation study on JAAD$_{beh}$ dataset and JAAD$_{all}$, respectively. Different model variants are denoted by Ours1, Ours2, ..., Ours7. By comparing Ours5 with Ours4 and Ours1 with the PCPA model, it shows that introducing global context can improve the model performance. If we further compare Ours4 with the PCPA model, it shows that using 2D CNN plus RNN instead of 3D CNN for visual feature encoding also has the advantage of extracting spatio-temporal features, hencing improving model performance. In terms of fusion strategies, the proposed hybrid fusion strategy achieved the best performance, which can be identified by comparing Ours with Ours5, Ours6, and Ours7.

    \begin{table*}[htb]
      \centering
      \centering
      \caption{Ablation Study on JAAD Behavior Subset}
      \begin{threeparttable}
      \small
      \renewcommand{\arraystretch}{1.3}
      
      \begin{tabular}{ccccccccc}
      
      \hline
      
      \hline
      
      \multirow{2}{*}{\bf Models}
      & \multicolumn{3}{|c|}{\begin{tabular}[c]{@{}c@{}} {\bf Model Variants} \end{tabular}} 
      & \multicolumn{5}{c}{\begin{tabular}[c]{@{}c@{}} {\bf JAAD$_{beh}$} \end{tabular}} 
      \\ \cline{2-9} 
    
      & \multicolumn{1}{|c|}{\begin{tabular}[c]{@{}c@{}} Visual Encoder \end{tabular}}
      & \multicolumn{1}{c|}{\begin{tabular}[c]{@{}c@{}} Global Context \end{tabular}} 
      & \multicolumn{1}{c|}{\begin{tabular}[c]{@{}c@{}} Fusion Approach\end{tabular}} 
      
      & \multicolumn{1}{c|}{\begin{tabular}[c]{@{}c@{}} Accuracy \end{tabular}} 
      & \multicolumn{1}{c|}{\begin{tabular}[c]{@{}c@{}} AUC \end{tabular}} 
      & \multicolumn{1}{c|}{\begin{tabular}[c]{@{}c@{}} F1 Score \end{tabular}} 
      & \multicolumn{1}{c|}{\begin{tabular}[c]{@{}c@{}} Precision \end{tabular}} 
      & \multicolumn{1}{c}{\begin{tabular}[c]{@{}c@{}} Recall \end{tabular}} 
      
       \\ \cline{1-9}

      Ours     &VGG + GRU &\cmark &hybrid-fusion &0.62 &0.54 &{\bf0.74} &0.65 &\bf0.85 
      \\ \hline 
      
      \textbf{Ablations} & && & & & & & \\
      
      \hspace{3mm} Ours1      &3D CNN &\cmark &later-fusion &0.59 &0.53 &0.69 &0.65 &0.75 
      \\
      \hspace{3mm} Ours2      &3D CNN &\cmark &early-fusion &0.59 &0.54 &0.69 &0.65 &0.74 
      \\
      \hspace{3mm} Ours3      &3D CNN &\cmark &hierarchical-fusion &0.57 &0.48 &0.70 &0.62 &0.81 
      \\  
      \hspace{3mm} Ours4      &VGG + GRU &\xmark &later-fusion &0.59 &0.51 &0.72 &0.63 &0.83 
      \\
      \hspace{3mm} Ours5      &VGG + GRU &\cmark &later-fusion &{\bf0.64} &{\bf0.59} &0.73 &\bf 0.68 &0.78 
      \\
      \hspace{3mm} Ours6      &VGG + GRU &\cmark &early-fusion &0.60 &0.56 &0.70 &0.67 &0.73 
      \\
      \hspace{3mm} Ours7      &VGG + GRU &\cmark &hierarchical-fusion &0.54 &0.50 &0.64 &0.63 &0.65 
      \\\hline\hline

      \hline
      
      \end{tabular}

      \begin{tablenotes}
              \footnotesize
              \item[$\bullet$] The {\bf bold} result means the best in the models. 
            \end{tablenotes}
          \end{threeparttable}
      
      \label{table:jad_beh_ablation}
      \end{table*}

      \begin{table*}[htb]
      \centering
      \centering
      \caption{Ablation Study on JAAD All Dataset}
      \begin{threeparttable}
      \small
      \renewcommand{\arraystretch}{1.3}
      
      \begin{tabular}{ccccccccc}
      
      \hline
      
      \hline
      
      \multirow{2}{*}{\bf Models}
      & \multicolumn{3}{|c|}{\begin{tabular}[c]{@{}c@{}} {\bf Model Variants} \end{tabular}} 
      & \multicolumn{5}{c}{\begin{tabular}[c]{@{}c@{}} {\bf JAAD$_{all}$} \end{tabular}} 
      \\ \cline{2-9} 
    
      & \multicolumn{1}{|c|}{\begin{tabular}[c]{@{}c@{}} Visual Encoder \end{tabular}}
      & \multicolumn{1}{c|}{\begin{tabular}[c]{@{}c@{}} Global Context \end{tabular}} 
      & \multicolumn{1}{c|}{\begin{tabular}[c]{@{}c@{}} Fusion Approach\end{tabular}} 
      
      & \multicolumn{1}{c|}{\begin{tabular}[c]{@{}c@{}} Accuracy \end{tabular}} 
      & \multicolumn{1}{c|}{\begin{tabular}[c]{@{}c@{}} AUC \end{tabular}} 
      & \multicolumn{1}{c|}{\begin{tabular}[c]{@{}c@{}} F1 Score \end{tabular}} 
      & \multicolumn{1}{c|}{\begin{tabular}[c]{@{}c@{}} Precision \end{tabular}} 
      & \multicolumn{1}{c}{\begin{tabular}[c]{@{}c@{}} Recall \end{tabular}}

       \\ \cline{1-9}

      Ours     &VGG + GRU &\cmark &hybrid-fusion &\bf0.83 &\bf0.82 &\bf0.63 &\bf0.51 &0.81  \\ \hline 
      
      \textbf{Ablations} & & & & & & & &  \\
      
      \hspace{3mm} Ours1      &3D CNN &\cmark &later-fusion  &0.77 &0.77 &0.54 &0.42 &0.76  
      \\
      \hspace{3mm} Ours2      &3D CNN &\cmark &early-fusion  &0.77 &0.74 &0.51 &0.41 &0.69  
      \\
      \hspace{3mm} Ours3      &3D CNN &\cmark &hierarchical-fusion &0.78 &0.77 &0.55 &0.43 &0.75 
      \\  
      \hspace{3mm} Ours4      &VGG + GRU &\xmark &later-fusion  &0.75 &0.79 &0.54 &0.40 &\bf 0.85
      \\
      \hspace{3mm} Ours5      &VGG + GRU &\cmark &later-fusion  &0.77 &0.80 &0.56 &0.43 &0.84
      \\
      \hspace{3mm} Ours6      &VGG + GRU &\cmark &early-fusion &0.79 &0.74 &0.52 &0.43 &0.66
      \\
      \hspace{3mm} Ours7      &VGG + GRU &\cmark &hierarchical-fusion &0.80 &0.81 &0.59 &0.46 &0.84  
      \\\hline\hline

      \hline
      
      \end{tabular}

      \begin{tablenotes}
              \footnotesize
              \item[$\bullet$] The {\bf bold} result means the best in the models. 
            \end{tablenotes}
          \end{threeparttable}
      
      \label{table:jad_all_ablation}
      \end{table*}

\subsection{Qualitative Results}

Figure~\ref{fig:results} provides qualitative results for the proposed model of pedestrian crossing intention prediction. We mainly compared the proposed method with the PCPA model. In the provided examples, our method correctly predicted the crossing intention but the PCPA failed. Taking a closer look at the examples, the following argument is raised. Without utilizing the global context, the task of crossing intention prediction may face the problems of (1) unknown direction of the pedestrian (Case a in Figure~\ref{fig:results}), (2) occlusion (Case b in Figure~\ref{fig:results}), and (3) poor vision (Case c in Figure~\ref{fig:results}). Global context can provide additional information to account for the interaction between the whole scene and the target pedestrian. 

\begin{figure*}
  \includegraphics[width=\linewidth]{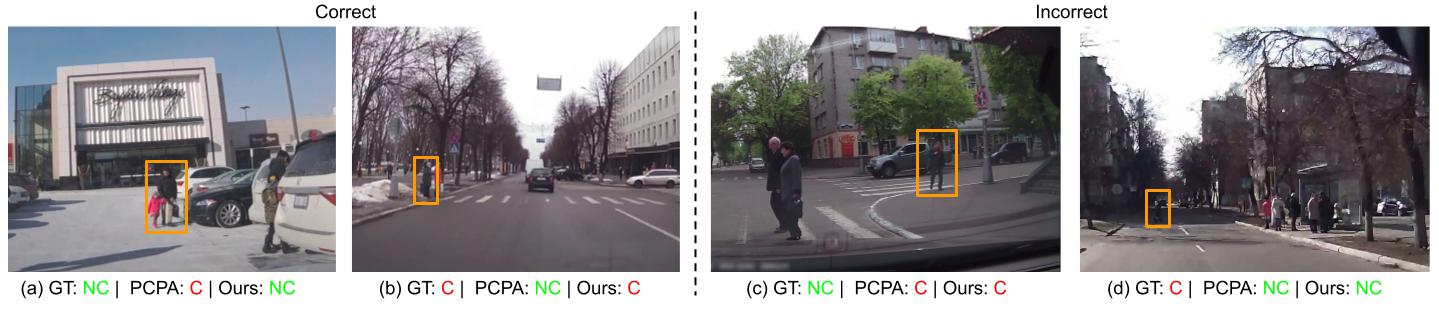}
  \caption{More qualitative results. (a) and (b) show the cases of correct predictions by the proposed model but the PCPA failed. (c) and (d) show the results when both the proposed and the PCPA model failed.}
  \label{fig:results_more_qualitative}
\end{figure*}

Figure~\ref{fig:results_more_qualitative} provides more qualitative results to analyze the advantages of the proposed model over the PCPA model as well as a few of failure cases. Figure~\ref{fig:results_more_qualitative}-(a) and Figure~\ref{fig:results_more_qualitative}-(b) show the cases when the proposed model generated correct predictions but the PCPA failed. The main reason is that our model considers the global visual context that contains the semantic segmentation of the drivable area. The model can learn from this whether the pedestrian is moving toward or on the drivable area, which is an important indicator of pedestrian crossing intention. 

Figure~\ref{fig:results_more_qualitative}-(c) and Figure~\ref{fig:results_more_qualitative}-(d) show the cases when both the proposed model and the PCPA failed. Figure~\ref{fig:results_more_qualitative}-(c) shows an intersection scenario. The pedestrian (yellow bounding box) has already crossed the ego road but near the edge of the road on the other side. This may mislead the model to generate a prediction of crossing. The failure in figure~\ref{fig:results_more_qualitative}-(d) was mainly due to the poor illumination such that the model cannot obtain enough detailed feature. 




\section{Conclusion}

In this work, we proposed a novel method for vision-based pedestrian crossing intention prediction. Our method explicitly considers the global context as a channel representing the interaction between the target pedestrian and the whole scene. We also proposed a hybrid fusion strategy for different features using 2D CNNs, RNNs, and attention mechanisms. Experiments on the JAAD dataset show that the proposed method achieves the state-of-the-art against baseline methods in the pedestrian action prediction benchmark.

Future work can focus on improving our model's robustness in unexpected situations, e.g., poor vision and occlusion. Additionally, feature fusion with more information sources can be explored. Finally, fine-tuning the model for particular pedestrian subsets, such as children and disabled people, can increase overall safety and performance.

\section*{Acknowledgment}

Material reported here was supported by the United States Department of Transportation under Award Number 69A3551747111 for the Mobility21 University Transportation Center. Any findings, conclusions, or recommendations expressed herein are those of the authors and do not necessarily reflect the views of the sponsors.

\ifCLASSOPTIONcaptionsoff
  \newpage
\fi



\bibliographystyle{IEEEtran}
\bibliography{IEEEabrv, mybib}

%



%


\begin{IEEEbiography}[{\includegraphics[width=1in,height=1.25in,clip,keepaspectratio]{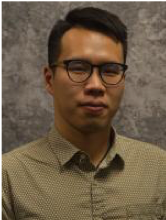}}]{Dongfang Yang} (Member, IEEE)
  received his bachelor's degree in microelectronics from Sun Yat-sen University, Guangzhou, China, in 2014. He had been with The Ohio State University since 2015 and received his Ph.D. in Electrical and Computer Engineering from The Ohio State University, Columbus, OH, United States., in 2020. 
  
  He is currently a senior algorithm engineer at Changan Automobile and a postdoc researcher at Chongqing University in Chongqing, China. His research interests include data analysis, machine learning, deep learning, and control systems, with applications in behavior prediction, decision-making, and motion planning in autonomous systems.
  \end{IEEEbiography}

\begin{IEEEbiography}[{\includegraphics[width=1in,height=1.25in,clip,keepaspectratio]{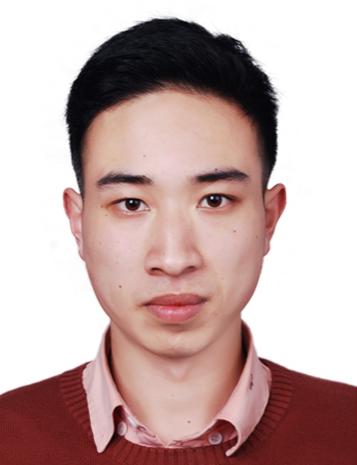}}]{Haolin Zhang}
  received the B.E. degree from Shanghai University of Electric Power, China, and the M.S. degree from The Ohio State University, USA. His research interests are artificial intelligence, computer vision, and perception in the applications of intelligent vehicles.
  \end{IEEEbiography}

\begin{IEEEbiography}[{\includegraphics[width=1in,height=1.25in,clip,keepaspectratio]{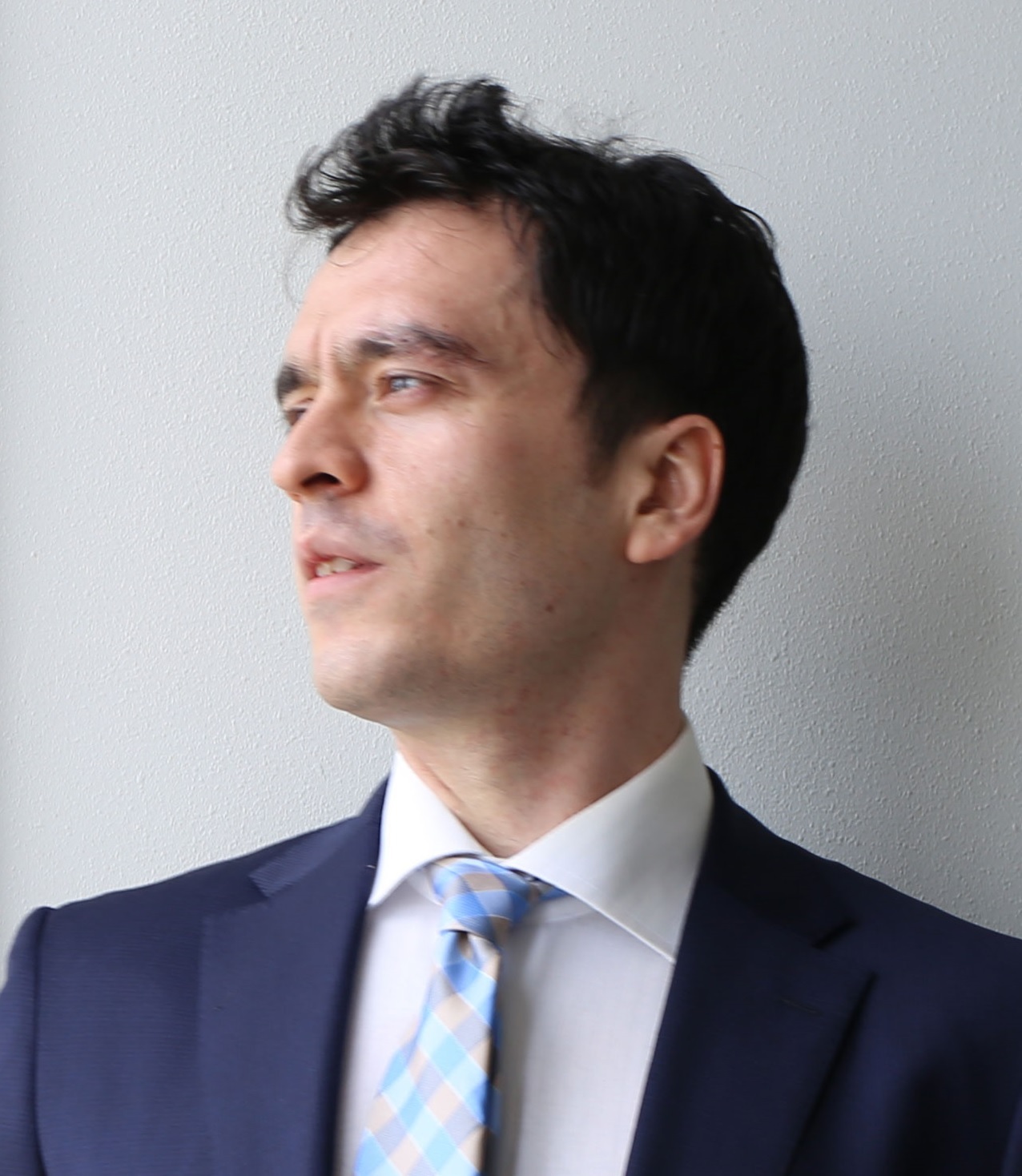}}]{Ekim Yurtsever} (Member, IEEE)
  received his B.S. and M.S. degrees from Istanbul Technical University in 2012 and 2014 respectively. He received his Ph.D. in Information Science in 2019 from Nagoya University, Japan. Since 2019, he has been with the Department of Electrical and Computer Engineering, The Ohio State University as a research associate.
      
  His research focuses on artificial intelligence, machine learning, computer vision, reinforcement learning, intelligent transportation systems, and automated driving systems.
  \end{IEEEbiography}

\begin{IEEEbiography}[{\includegraphics[width=1in,height=1.25in,clip,keepaspectratio]{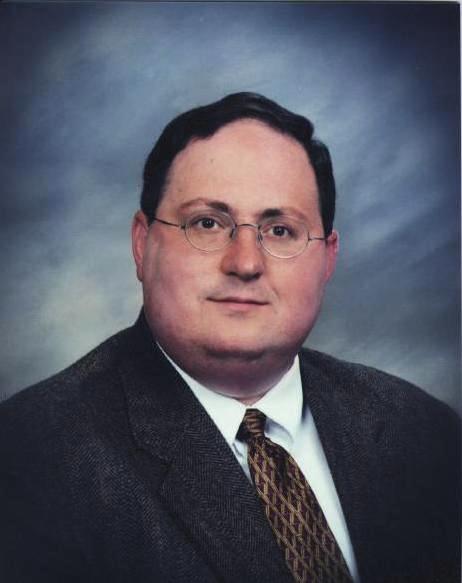}}]{Keith A. Redmill}
  (S’89–M’98–SM’11) received the B.S.E.E. and B.A. degrees in mathematics from Duke University, Durham, NC, USA, in 1989 and the M.S. and Ph.D. degrees from The Ohio State University, Columbus, OH, USA, in 1991 and 1998, respectively. Since 1998, he has been with the Department of Electrical and Computer Engineering, The Ohio State University, initially as a Research Scientist. He is currently a Research Associate Professor.
  
  He is a coauthor of the book \textit{Autonomous Ground Vehicles}. His research interests include autonomous vehicles and robots, intelligent transportation systems, vehicle and bus tracking, wireless data communication, cellular digital packet data, Global Positioning System and Geographic Information System technologies, large hierarchical systems, real-time and embedded systems, hybrid systems, control theory, dynamical systems theory, cognitive science, numerical analysis and scientific computation, and computer engineering.
  \end{IEEEbiography}

\begin{IEEEbiography}[{\includegraphics[width=1in,height=1.25in,clip,keepaspectratio]{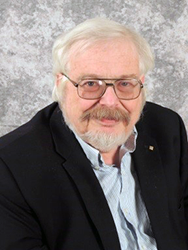}}]{Ümit Özgüner} (S’72–M’75–F’10) 
  Prof. Emeritus Ümit Özgüner, TRC Inc. Chair on ITS at The Ohio State University, is a well know expert on Intelligent Vehicles. He holds the title of “Fellow” in IEEE for his contributions to the theory and practice of autonomous ground vehicles and is the Editor in Chief of the IEEE ITS Society, Transactions on Intelligent Vehicles.

  He has led and participated in many autonomous ground vehicle related programs like DoT FHWA Demo’97, DARPA Grand Challenges and the DARPA Urban Challenge. His research has been (and is) supported by many industries including Ford, GM, Honda and Renault. He has published extensively on control design and vehicle autonomy and has co-authored a book on Ground Vehicle Autonomy. His present projects are on Machine Learning for driving, pedestrian modeling at OSU and participates externally on V\&V and risk mitigation, and self-driving operation of specialized vehicles. Professor Ozguner has developed and taught a course on Ground Vehicle Autonomy for many years and has advised over 35 students during their studies towards a PhD.
  \end{IEEEbiography}




\end{document}